\documentclass{llncs}
\usepackage{a4,a4wide}
\usepackage[english]{babel}

\usepackage{listings}
\usepackage{xspace}

\newcommand{\idp}{{\sc IDP}\xspace}
\newcommand{\gidl}{{\sc GidL}\xspace}
\newcommand{\minisatid}{{\sc MinisatID}\xspace}
\newcommand{\lua}{{\sc Lua}\xspace}
\newcommand{\tptp}{{\sc TPTP}\xspace}
\newcommand{\idpdraw}{{\sc IDPDraw}\xspace}
\newcommand{\enfragmo}{{\sc Enfragmo}\xspace}
\newcommand{\cplusplus}{{\tt C++}\xspace}
\newcommand{\dlv}{{\tt DLV}\xspace}
\newcommand{\clp}{{\tt CLP}\xspace}

\newcommand{\clingo}{{\sc clingo}\xspace}
\newcommand{\cplex}{{\sc CPLEX}\xspace}

\newcommand{\mozart}{{\sc Mozart}\xspace}
\newcommand{\comet}{{\sc Comet}\xspace}
\newcommand{\zinc}{{\sc Zinc}\xspace}
\newcommand{\prolog}{{\sc Prolog}\xspace}
\newcommand{\ourlanguage}{{\sc declimp}\xspace}

\title{A prototype of a knowledge-based programming environment}
\author{Stef De Pooter \and Johan Wittocx \and Marc Denecker}
\institute{Department of Computer Science, K.U.\ Leuven}
\date{}

\begin{document}
\maketitle

\begin{abstract}
In this paper we present a proposal for a knowledge-based programming environment. In such an environment, declarative background knowledge, procedures, and concrete data are represented in suitable languages and combined in a flexible manner. This leads to a highly declarative programming style. We illustrate our approach on an example and report about our prototype implementation.
\end{abstract}

\section{Context}

An obvious requirement for a powerful and flexible programming paradigm seems to be that within the paradigm different types of information can be expressed in suitable languages. However, most traditional programming paradigms and languages do not really have this property. In imperative languages, for example, non-executable background knowledge can not be described. The consequences become clear when we try to solve a scheduling problem in an imperative language: the background knowledge, the constraints that need to be satisfied by the schedule, gets mixed up with the algorithms. This makes adding new constraints and finding and modifying existing ones cumbersome.

On the other hand, most logic-based declarative programming paradigms lack the capacity to express procedures. Typically, they consist of a logic together with one specific type of inference. For example, Prolog uses Horn clause logic and does querying, in Description Logic the studied task is deduction, and Answer Set Programming and Constraint Programming make use of model generation. In such paradigms, whenever we try to perform a task that does not fit the inference mechanism at hand, the declarative aspect of the paradigm disappears. For example, when we try to solve a scheduling problem (which is a typical model-generation problem) in Prolog, then we need to represent the schedule as a term, say a list (rather than as a logical structure), and as a result the constraints do not really reside in the logic program, but will have to be expressed by clauses that iterate over a list~\cite{ccai/DeneckerDW90}. Proving that a certain requirement is implied by another, is possible (in theory) for a theorem prover, but not in ASP. Etc.

To overcome these restrictions of existing paradigms, we propose a paradigm in which each component can be expressed in an appropriate language. We distinguish three components: procedures, (non-executable) background knowledge, and concrete data. For the first we need an imperative language, for the second an (expressive) logic, for the third a logical structure (which corresponds to a database). The connection between these components is mostly realized by various reasoning tasks, such as theorem proving, model generation, model checking, model revision, belief revision, constraint propagation, querying, datamining, visualization, etc.

The idea to support multiple forms of inference for the same logic or even for the same theories, was argued in \cite{iclp/DeneckerV08}. Here it is argued that logic has a more flexible, multifunctional and therefore also more declarative role for problem solving than provided in many declarative programming paradigms, where typically one form of inference is central and theories are written to be used for this form of inference, sometimes even for a specific algorithm implementing this form of inference (such as \prolog resolution). This view was therefore called the Knowledge Base System paradigm for declarative problem solving. The framework presented here is based on this view and goes beyond it in the sense that it offers a programming environment in which complex tasks can be programmed using multiple forms of inference and processing tools.

\section{Overview of the language and system}

To try out the above mentioned ideas in practice, we built a prototype interpreter that supports some basic reasoning tasks and a set of processing tools on high-level data such as vocabularies, structures and theories. In this section we will highlight various decisions in the design of our programming language and interpreter. In the next section we will illustrate the usage of the language with an example. We named our language \ourlanguage, which is an aggregation of ``declarative'' and ``imperative''.

\subsection{Program structure}

A \ourlanguage program typically contains several blocks of code. Each block is either a procedure, a vocabulary (which is a list of sort, predicate and function names), a logic theory over vocabularies (which describes a piece of background knowledge using the relation and function names of its vocabulary), or a (possibly three-valued) structure over vocabularies. The latter represent databases over their vocabularies. To bring more structure into a program and to be able to work with multiple files, namespaces and include statements are provided.

Because vocabularies, logic theories and databases are not executable, and a program needs to be executed, control of a \ourlanguage program is always in the hands of the procedures. Moreover, when a {\tt main} procedure is available, the run of the program will start with the execution of this procedure. When there is no {\tt main} procedure, the user can run commands in an interactive shell, after parsing the program.

In the next sections, we will describe the languages for the respective components in a \ourlanguage program.

\subsection{Knowledge representation language}
For representing background knowledge we use an extended version of classical logic. A first advantage in using this language lies in the fact that classical logic is the best known and most studied logic. Also, classical logic has the important property that its informal semantics corresponds to its formal semantics. In other words, in classical logic the meaning of expressions\footnote{We mean expressions that occur in practice, not artificially constructed sentences that do not really have meaning in real life.} is intuitively clear. This is an important requirement in the design of a language that is accessible to a wider audience. Furthermore, there are already numerous declarative systems that use a language based on classical logic, or can easily be translated to it. Think of the languages of most theorem provers, various Description logics, and the language of model generators such as \idp~\cite{lash08/WittocxMD08,url:idp} and \enfragmo~\cite{MitchellTHM06}.

Research in the Knowledge Representation and Reasoning community has clearly shown that classical logic is in many ways insufficient. Aggregates and (recursive) definitions are well-known concepts that are common in the background knowledge of many applications, and which can generally not, or not in a concise and intuitively clear manner, be expressed in first-order logic. In \ourlanguage we use an order-sorted version of first-order logic, extended with inductive definitions~\cite{tocl/DeneckerT08}, aggregates~\cite{tplp/PelovDB07}, (partial) functions and arithmetic.

\subsection{Structures}

Structures in \ourlanguage are written in a simple language that allows to enumerate all elements that belong to a sort and all tuples that belong to a relation or function. As an alternative to enumerating a relation, it is also possible to specify the relation in a procedural way, namely as all the tuples for which a given procedure returns `true'. Furthermore, the interpretation of a function can be specified by a procedure, somewhat similar to ``external procedure'' in \dlv~\cite{lncs/CalimeriCIL08}.

As mentioned before, structures in \ourlanguage are not necessarily two-valued. Three-valued structures are useful for representing incomplete information (which might be completed during the run of the program). To enumerate a three-valued relation (or function), two out of three of the following sets must be provided: tuples that certainly belong to the relation, tuples that certainly do not belong to the relation, and tuples for which it is unknown whether they belong to the relation or not. The third set can always be computed from the two given sets.

\subsection{Procedures}

The imperative programming language in our prototype system is \lua~\cite{SPE/IerusalimschyFC96}. The main reason for this choice is the fact that \lua is a lightweight scripting language and also because it has a good \cplusplus API~\cite{queue/IerusalimschyFC11}. This facilitates on the one hand the compilation of programs written in \ourlanguage and, on the other hand, the integration with the components of our \ourlanguage interpreter, which is written in \cplusplus. When we do not take those reasons into account, any other imperative language is candidate.

In procedures, various reasoning methods on theories and structures can be called. Currently, the most important tasks supported by the \ourlanguage-interpreter are the following:

\begin{description}
\item[Finite model expansion:] Given a three-valued structure $S$ and a theory $T$, find a completion of $S$ to a two-valued structure that satisfies $T$. This is essentially a generalization of the reasoning task performed by ASP solvers, constraint programming systems, Alloy analyzers, etc. It is suitable for problems such as scheduling, planning and diagnosis. In our \ourlanguage interpreter, model expansion is implemented by calls to the \idp system~\cite{lash08/WittocxMD08}, which consists of the grounder \gidl~\cite{jair/WittocxMD10} and solver \minisatid~\cite{sat/MarienWDB08}.

\item[Finite model checking:] Check whether a given two-valued structure is a model of a theory. This is an instance of model expansion and is implemented as such.

\item[Constraint propagation:] Deduce facts that must hold in all models of a given theory which complete a given three-valued structure. This is a useful mechanism in configuration systems~\cite{ppdp/VlaeminckVD09} and for query answering in incomplete databases~\cite{tods/DeneckerCBA10}. The propagation algorithm we implemented is described in~\cite{CoRR/wittocxDB11}.

\item[Querying:] Given an FO formula $\varphi$ and a two-valued structure $S$, find all substitutions for free variables of $\varphi$ that make $\varphi$ true in $S$. The implementation of this mechanism makes use of Binary Decision Diagrams as described in ~\cite{jair/WittocxMD10}.

\item[Theorem proving:] Given two FO theories $T_1$ and $T_2$, check whether $T_1 \models T_2$. This is implemented by calling a theorem prover provided by the user. In principle, any theorem prover that accepts \tptp~\cite{Sutcliffe09} can be used.

\item[Visualization:] Show a visual representation of a given structure. We implement this by calling \idpdraw, a tool for visualizing finite structures in which visual output is specified declaratively by definitions in our knowledge representation language or in ASP.
\end{description}
The values returned by the reasoning methods can be used in other reasoning methods and \lua-statements. We will illustrate this with an example in the next section. 

\section{Programming in \ourlanguage}

Say we want to write an application that allows players to solve sudoku puzzles. Such an application should be able to perform tasks such as generating puzzles, showing puzzles on the screen, checking whether solutions (player's choices) satisfy the sudoku rules, giving hints to the player, etc. In this application the different components we described before can clearly be distinguished: 
(1) the background knowledge consists of a logic theory containing the well-known sudoku constraints; 
\[
\begin{array}{l}
	\forall r \forall n \exists! c : Sudoku(r,c) = n\\
	\forall c \forall n \exists! r : Sudoku(r,c) = n\\
	\forall b \forall n \exists! r \exists! c : InBlock(b,r,c) \wedge Sudoku(r,c) = n\\
	\forall b \forall r \forall c : InBlock(b,r,c) \Leftrightarrow b = ((r-1)/3)*3 + ((c-1)/3) + 1\\
\end{array}
\]
(2) the data is stored in logical structures representing puzzles, and (partial and complete) solutions; and 
(3) the tasks we want it to perform, can be implemented using well-known inference methods.

Below we show (part of) a \ourlanguage program. This code shows the use of an include statement and a namespace, and the declaration of a vocabulary {\tt sudokuVoc} and a theory {\tt sudokuTheory}, where the latter is simply an ASCII version of the theory shown above. Also note the {\tt main} procedure at the bottom, which will be called when this program is passed to the interpreter.

\begin{lstlisting}
#include "grid.idp"

namespace sudoku {

	vocabulary sudokuVoc {
		extern vocabulary grid::simpleGridVoc
		type Num isa nat
		type Block isa nat
		Sudoku(Row,Col) : Num
		InBlock(Block,Row,Col)
	}
	
	theory sudokuTheory : sudokuVoc {
		! r n : ?1 c : Sudoku(r,c) = n.
		! c n : ?1 r : Sudoku(r,c) = n.
		! b n : ?1 r c : InBlock(b,r,c) & Sudoku(r,c) = n.
		! r c b : InBlock(b,r,c) <=> b = ((r-1)/3)*3 + ((c-1)/3) + 1.
	}
	
	procedure solve(input) {
		return modelExpand(sudokuTheory,input)
	}

	procedure printSudoku(puzzle) {
		-- code for visualizing a sudoku puzzle.
	}

	procedure createSudoku() {
		math.randomseed(os.time())
		local puzzle = grid::makeEmptyGrid(9) -- defined in grid.idp
	
		stdoptions.nrmodels = 2	
		local currsols = modelExpand(sudokuTheory,puzzle)
		while #currsols > 1 do
			repeat
				col = math.random(1,9)
				row = math.random(1,9)
				num = currsols[1][sudokuVoc::Sudoku](row,col)
			until num ~= currsols[2][sudokuVoc::Sudoku](row,col)
			makeTrue(puzzle[sudokuVoc::Sudoku].graph,{row,col,num})
			currsols = modelExpand(sudokuTheory,puzzle)
		end 
	
		printSudoku(puzzle)
	}
}

procedure main() {
	sudoku::createSudoku()
}
\end{lstlisting}

\noindent Let us have a closer look at procedure {\tt createSudoku} for creating sudoku puzzles. First it initializes an empty puzzle by instantiating a new logical structure. This is done by calling a procedure {\tt makeSquareGrid} which instantiates a structure with data about a generic grid of a certain size, and then adding domains for numbers and blocks particular for sudoku grids. 

The second part of the procedure adds numbers to the grid until there is only one solution left for the puzzle. This is realized by performing model expansion (by calling {\tt modelExpand}) to find two models of the theory that extend the given partially filled in puzzle. When two models are found, the algorithm selects a number that is unique for the first solution (that is, the number at the same position in the second solution is different) and is not yet present in the puzzle. When such an entry is found, it is added to the puzzle by making the tuple {\tt \{row,col,num\}} true in the interpretation of the function {\tt Sudoku(Row,Col):Num}. Next, the procedure ask for two new models, and the process starts over. When only one model is found, the iteration stops, and procedure {\tt printSudoku} is called to show the result on the screen using the visualization tool mentioned in the previous section.

\section{Related work}

There have been many proposals in the literature to combine procedural and declarative languages. A frequently occuring combination is that of a procedural language in which a program can post constraints expressed in an (often ad-hoc) declarative constraint language, while other primitives allow to call the constraint-solving process on the constraint store, express heuristics or call other processes, for example to edit or visualize output. Examples of systems with such languages are \cplex~\cite{url:cplex}, \mozart~\cite{moz/VanRoy05} and \comet~\cite{cp/MichelH05}. These systems differ from \ourlanguage in the sense that they offer only one kind of inference, namely constraint solving. A similar remark can be made about \clp and \prolog systems with support for constraint propagation. Here the ``procedural language'' is the \prolog language under its procedural semantics. In our system high-level concepts such as vocabularies, theories and structures are treated as first-class citizens that can be operated upon by arbitrary inference and processing tools, which offers more flexibility.

For another group of systems, control over execution of programs is in hands of one inference mechanism -- or at least that inference is the main mechanism -- and an integrated procedural language then allows users to stear some aspects of the inference mechanism, or for example format input and output, but do not allow to take over control. Examples of such systems are \clingo~\cite{url:clingo} and \zinc~\cite{constraints/MarriottNRSBW08}. The procedural languages in these systems have a more limited task then the one in \ourlanguage. In \ourlanguage the procedures are in control during execution, not just one of the inference mechanisms.

\section{Conclusion}

We have presented a knowledge-based programming environment, providing a declarative language for expressing background knowledge, an imperative programming language for writing procedures, and logical structures for expressing concrete data. The system also provides some state-of-the-art inference tools for performing various reasoning tasks. 

We believe that a programming environment like the one proposed here overcomes some of the limitations of ``single-programming-style'' paradigms, by allowing a programmer to express the different types of information in software applications in appropriate languages. Making this explicit distinction in different types of information will increase readability, maintainability and reusability of programming code.

\bibliographystyle{plain}


\end{document}